\documentclass[12pt]{article}
\usepackage[utf8]{inputenc}
\usepackage[margin=1in]{geometry}

\usepackage{graphicx}
\usepackage{cite}
\usepackage{framed}
\usepackage{xcolor}
\usepackage{multicol}
\usepackage[framemethod=tikz]{mdframed}
\usepackage{amsmath}    
\usepackage{amssymb}    
\usepackage{amsfonts}   
\usepackage{comment}
\usepackage{booktabs}
\usepackage{amsmath}
\usepackage{graphicx}
\usepackage[table]{xcolor}
\usepackage{fontspec}
\usepackage[super]{natbib}
\usepackage[colorlinks=true, citecolor=blue, linkcolor=blue]{hyperref}

\newmdenv[
  backgroundcolor=gray!15,
  linecolor=black,
  linewidth=1pt,
  roundcorner=5pt
]{myshadedbox}

\colorlet{shadecolor}{gray!15}

\begin{document}

\begingroup
\renewcommand{\thefootnote}{\fnsymbol{footnote}}

\endgroup

\begin{quote}
\begin{center}
    {
    \vspace{1em}
    
    \rule{0.8\textwidth}{4pt}
    \vspace{1em}
    
    \Large \bf A Hierarchical Benchmark of Foundation Models for Dermatology}\\[1em]
    
    \rule{0.8\textwidth}{2pt}  
    \vspace{1em}

    \normalsize
    \textbf{Furkan Yuceyalcin}$^{1,*}$,
    \textbf{Abdurrahim Yilmaz}$^{2,*,\dagger}$,
    \textbf{Burak Temelkuran}$^{2}$\\[0.5em]

    {\footnotesize
    \textsuperscript{1}Yildiz Technical University
    \textsuperscript{2}Imperial College London
    }\\[1em]


\vspace{1em}

\noindent

\textbf{Abstract}

\end{center}
Foundation models have transformed medical image analysis by providing robust feature representations that reduce the need for large-scale task-specific training. However, current benchmarks in dermatology often reduce the complex diagnostic taxonomy to flat, binary classification tasks, such as distinguishing melanoma from benign nevi. This oversimplification obscures a model's ability to perform fine-grained differential diagnoses, which is critical for clinical workflow integration. This study evaluates the utility of embeddings derived from ten foundation models, spanning general computer vision, general medical imaging, and dermatology-specific domains, for hierarchical skin lesion classification. Using the DERM12345 dataset, which comprises 40 lesion subclasses, we calculated frozen embeddings and trained lightweight adapter models using a five-fold cross-validation. We introduce a hierarchical evaluation framework that assesses performance across four levels of clinical granularity: 40 Subclasses, 15 Main Classes, 2 and 4 Superclasses, and Binary Malignancy. Our results reveal a "granularity gap" in model capabilities: MedImageInsights achieved the strongest overall performance (97.52\% weighted F1-Score on Binary Malignancy detection) but declined to 65.50\% on fine-grained 40-class subtype classification. Conversely, MedSigLip (69.79\%) and dermatology-specific models (Derm Foundation and MONET) excelled at fine-grained 40-class subtype discrimination while achieving lower overall performance than MedImageInsights on broader classification tasks. Our findings suggest that while general medical foundation models are highly effective for high-level screening, specialized modeling strategies are necessary for the granular distinctions required in diagnostic support systems.
\end{quote}

{\small $^{\dagger}$Corresponding author: a.yilmaz23@imperial.ac.uk}
{\small $^{*}$These authors contributed equally to this work.}

\newpage

\section{Introduction}
Dermatological diagnosis is an inherently hierarchical process\cite{Braun2002}. When a clinician evaluates a skin lesion, the cognitive workflow proceeds from broad categorization (e.g., melanocytic versus non-melanocytic), to risk assessment of malignancy, and finally to specific diagnosis. A robust diagnostic system must not only distinguish benign from malignant lesions but also differentiate between visually similar but biologically distinct entities, such as a compound nevus and a dermal nevus. 

The automation of skin lesion analysis historically relied on Convolutional Neural Networks (CNNs) trained end-to-end on specific datasets\cite{goyal_2020, Milton2019AutomatedSL}. While effective, this paradigm is resource-intensive and often yields models that generalize poorly across different clinical settings or imaging devices. Recently, foundation models have emerged as a powerful alternative\cite{Paschali_2025, Liu2024, woerner_and_baumgartner, embeddings_to_acc_Li2025}. By pretraining on vast repositories of natural or medical images using self-supervised learning or image-text alignment, these models learn rich, transferable feature representations. In the general medical domain, models such as BiomedCLIP\cite{Zhang_biomedclip}, MedImageInsights \cite{Codella_2024_medimageinsights}, and MedSigLip\cite{sellergren2025medgemmatechnicalreport} have demonstrated that leveraging diverse biomedical data can yield robust performance across various modalities, from radiology to histopathology\cite{Neidlinger2025_path_bench, lee2025, wang2023}. Simultaneously, dermatology-specific foundation models, including PanDerm\cite{Yan2025_panderm}, SkinVL\cite{mmskin} and Derm Foundation \cite{GoogleDermFoundation}, have been developed to capture the unique visual semantics of skin diseases.

Despite the rapid increase of foundation models in healthcare, a critical gap remains in their evaluation. Existing benchmarks typically focus on flat classification metrics or binary screening accuracy\cite{Xu2024.04.17.24305983, jin_fairfedfm, wang2023, huix2024}. Similarly, earlier benchmarks \cite{Milton2019AutomatedSL, daneshjou_disparities_in_dermai} and recent few-shot evaluations \cite{woerner_and_baumgartner} often reduce the diagnostic task to binary melanoma detection or the multi class schema. Barata et al.\cite{barata_2019} argued for the necessity of hierarchical diagnosis and Yan et al.\cite{yan2025derm1mmillionscalevisionlanguagedataset} recently introduced hierarchical ontologies in the Derm1M dataset. However, a comparative benchmark of modern foundation model embeddings across the full diagnostic taxonomy is needed to understand their performance. Current evaluations fail to capture the 'granularity gap,' where a model might excel at high-level screening due to strong semantic priors but struggle to capture the subtle, intra-class visual variance required to distinguish biologically related but distinct entities.

In this study, we present a comprehensive hierarchical benchmark of foundation model embeddings for dermatology by using DERM12345 dataset \cite{yilmaz2024derm12345}. We systematically evaluate ten models categorized into three domains: general computer vision models (DINOv2\cite{oquab2024dinov2learningrobustvisual}, DINOv3\cite{siméoni2025dinov3}, CLIP\cite{Radford2021LearningTV_CLIP}, ResNet-50\cite{resnet50_2016}), general medical foundation models (MedSigLip\cite{sellergren2025medgemmatechnicalreport}, BiomedCLIP\cite{Zhang_biomedclip}, MedImageInsights\cite{Codella_2024_medimageinsights}), and dermatology-specific models (MONET\cite{Kim2024_MONET_2024}, PanDerm\cite{Yan2025_panderm}, Derm Foundation\cite{GoogleDermFoundation}). Instead of fine-tuning heavy backbones, we employ a standardized feature extraction pipeline followed by the training of lightweight adapters. This approach isolates the quality of the learned representation from the capacity of the downstream classifier. Our evaluation strategy moves beyond standard metrics by mapping model predictions to four levels of the diagnostic hierarchy: 40 Subclasses, 15 Main Classes, 2 and 4 Superclasses, and a Binary Malignancy classification. This multi-level analysis reveals trade-offs between semantic understanding and visual discrimination, providing guidance for researchers selecting foundation models for specific clinical applications, from triage apps to expert-level decision support systems.

\section{Methods}

\begin{figure}[ht!]
    \centering
    \includegraphics[width=0.8\textwidth]{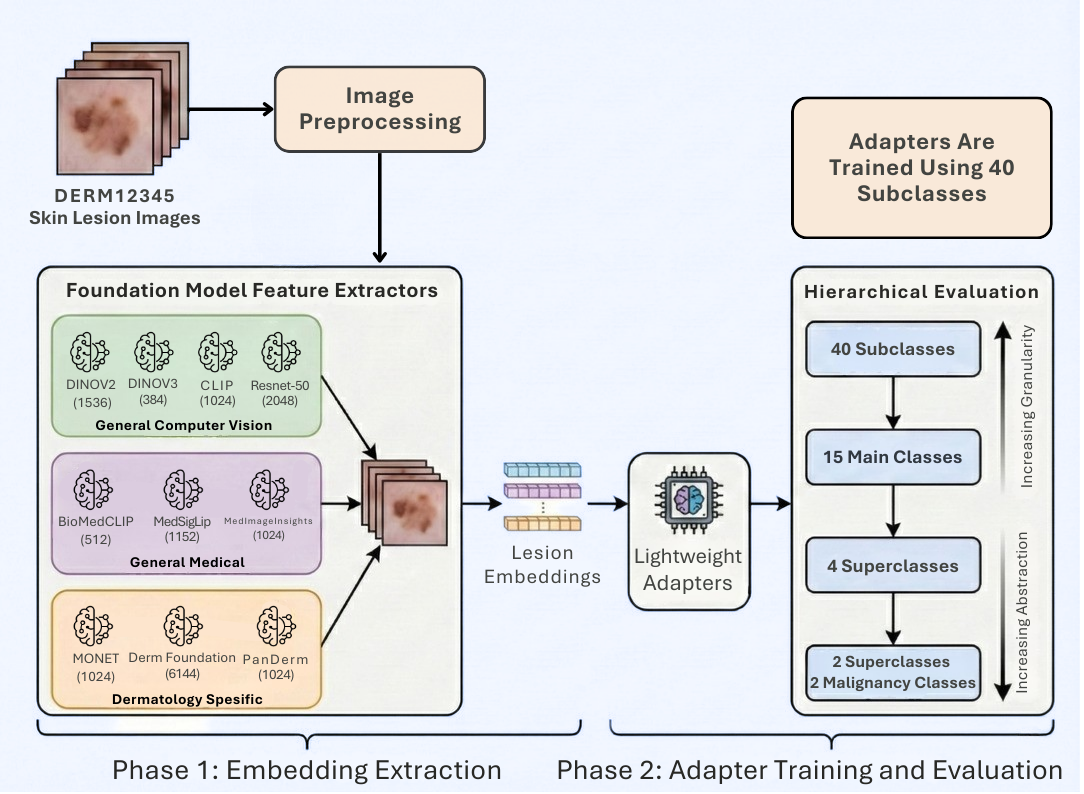}
    \caption{Overview of the experimental pipeline for benchmarking foundation models on the DERM12345 dataset \cite{yilmaz2024derm12345}. Phase 1: High-resolution skin lesion images are preprocessed and passed through ten frozen foundation model feature extractors, categorized into General Computer Vision (Green), General Medical (Purple), and Dermatology-Specific (Yellow) domains. Numbers in parentheses denote the embedding dimension size. Phase 2: The extracted embeddings are used to train lightweight adapter classifiers (K-Nearest Neighbors (KNN), Logistic Regression (LR), Support Vector Machines (SVM), Random Forest (RF), Multi-Layer Perceptron (MLP), and XGBoost). Performance is evaluated hierarchically by aggregating fine-grained predictions (40 Subclasses) into progressively coarser diagnostic levels (Main Classes, Superclasses, and Binary Malignancy), enabling an assessment of model performance across the full clinical taxonomy.}
    \label{fig:overview_pipeline}
\end{figure}

To evaluate the clinical utility of foundation model embeddings in dermatology, we designed a systematic three-phase experimental pipeline comprising data standardization, feature extraction, and hierarchical evaluation. The overall workflow, extending from raw image input to multi-level diagnostic prediction, is illustrated in Figure~\ref{fig:overview_pipeline}. This pipeline allows us to isolate the quality of the learned representations by keeping the foundation models frozen while training lightweight adapters. The following subsections detail the dataset characteristics, the selection of foundation models, the cross-validation protocol, and the hierarchical logic used to assess performance across the diagnostic taxonomy.

\subsection{Study Design and Dataset}
This study utilized the DERM12345 dataset \cite{yilmaz2024derm12345}, a large-scale collection of 12,345 dermatoscopic images representing a diverse range of skin lesions collected from multiple clinical sources in Türkiye. The dataset is characterized by a granular taxonomic structure that classifies lesions into 40 distinct subclasses. These subclasses are hierarchically organized into 15 main classes and 5 superclasses, providing a rich ground truth for evaluating model performance at varying levels of diagnostic abstraction. The dataset includes a pre-defined split of 9,860 images for training and 2,485 images for testing, partitioned based on unique patients to prevent data leakage between splits.

To facilitate the evaluation of malignancy detection, we mapped the dataset's native taxonomy to a binary malignancy label map. The original dataset taxonomy includes a category for "indeterminate" lesions, which represent actinic keratosis cases. Following conservative clinical protocols where such lesions are typically monitored rather than immediately excised, we grouped indeterminate lesions with the benign category for the binary classification task for simplification. Thus, this mapping reduced the original five superclasses defined in the dataset to four active superclasses for this benchmark: Melanocytic Benign, Melanocytic Malignant, Non-melanocytic Benign, and Non-melanocytic Malignant (Supplementary Table S1). We also evaluated an intermediate 2-class Superclass level distinguishing Melanocytic from Non-melanocytic lesions. This decision aligns the computational evaluation with the clinical goal of minimizing false negatives for clear malignancies while acknowledging the ambiguity of indeterminate cases.

\subsection{Foundation Models and Embedding Extraction}
We selected a diverse suite of foundation models to evaluate the impact of domain-specific pretraining on dermatological feature extraction. These models were categorized into three distinct domains. First, the General Vision category included DINOv2 (Base and Giant variants)\cite{oquab2024dinov2learningrobustvisual}, DINOv3\cite{siméoni2025dinov3}, and CLIP (Base and Large variants)\cite{Radford2021LearningTV_CLIP}. These models were pretrained on massive natural image datasets and serve as a baseline for general visual feature extraction capabilities. Second, the General Medical category included MedSigLip\cite{sellergren2025medgemmatechnicalreport}, BiomedCLIP\cite{Zhang_biomedclip}, and MedImageInsights\cite{Codella_2024_medimageinsights}. These models leverage large-scale medical image-text pairs or diverse radiological data, offering representations tuned to the broader medical domain. Third, the Dermatology Specific category included MONET\cite{Kim2024_MONET_2024}, PanDerm\cite{Yan2025_panderm}, and Derm Foundation\cite{GoogleDermFoundation}. These models were explicitly pretrained on dermatological images, theoretically offering the most semantically relevant features for skin lesion analysis.

We implemented a standardized inference pipeline to extract feature embeddings from these models. Each model was wrapped in a unified interface that handled model-specific preprocessing requirements, such as resizing and normalization. We utilized the models as fixed feature extractors without fine-tuning their weights. For each image in the dataset, the forward pass was computed, and the embedding vector was extracted from the model's final hidden state, typically via mean pooling or the specific classification token output as dictated by the model architecture. This process resulted in a standardized feature matrix for the training and test sets for each foundation model.

The extracted embedding dimensions varied significantly across the models (Supplementary Table S3). DINOv3 had the smallest size at 384, followed by BiomedCLIP at 512. The base versions of CLIP, DINOv2, and PanDerm utilized 768 dimensions. Several models, including CLIP-Large, MedImageInsights, MONET, and PanDerm-Large, output 1024 dimensions. Larger embeddings were produced by MedSigLip (1152) and DINOv2-Giant (1536), while the ResNet-50 baseline used 2048. Finally, Derm Foundation produced the largest embeddings with 6144 dimensions.

\subsection{Adapter Training and Optimization}
To assess the linear separability and utility of the extracted embeddings, we trained a set of lightweight adapter classifiers. We employed six distinct classification algorithms: K-Nearest Neighbors (KNN)\cite{knn_cover1967nearest}, Logistic Regression (LR)\cite{LR_cox1958regression}, Support Vector Machines (SVM)\cite{SVM_Cortes1995}, Random Forest (RF)\cite{RF_Breiman2001}, Multi-Layer Perceptron (MLP)\cite{MLP_rumelhart1986learning}, and XGBoost\cite{xgboost_2016}. Importantly, foundation models produce embeddings of vastly different dimensions, ranging from 384 in DINOv3 to 6144 in Derm Foundation. A single classification head might favor some models while unfairly punishing others. To ensure a fair comparison, we treat the adapter selection as a hyperparameter. For each foundation model, we report the peak performance achieved by the best-performing adapter. This approach isolates the quality of the embedding from the choice of classifier. It ensures we measure the full potential of each foundation model, rather than penalizing it for being incompatible with a specific algorithm.

We adopted a 5-fold stratified cross-validation strategy on the training set to ensure statistical robustness and prevent overfitting to a specific data split. For each fold, the training data was partitioned into training and validation subsets. Within each fold, we performed an hyperparameter search using GridSearchCV (Supplementary Table S2) to optimize model-specific parameters, such as the number of neighbors for KNN, the regularization strength C for SVM and LR, and the tree depth for RF and XGBoost. The metric for optimization was balanced accuracy, chosen to account for the natural class imbalance present in the dataset. The best-performing hyperparameters from the validation subset were then used to train the final model for that fold. This process yielded five optimized models for each combination of foundation model and classifier, effectively creating an ensemble of adapters for the final evaluation.

\subsection{Hierarchical Performance Evaluation}
We implemented a hierarchical evaluation framework to assess the trained adapters. Unlike standard benchmarks that report performance on a single label set, we evaluated the models on their ability to generalize across the clinical taxonomy. The adapters were trained solely on the fine-grained 40-class subclass labels. During the inference phase on the held-out test set, we generated probability distributions over these 40 classes.

To obtain predictions for coarser taxonomic levels, we aggregated the probabilities of the constituent subclasses according to the dataset's hierarchy. For example, the probability of the "Melanoma" main class was calculated as the sum of the probabilities of its subclasses, such as "Acral Nodular Melanoma" and "Lentigo Maligna Melanoma." This aggregation logic was applied to generate predictions for the 15 Main Classes, 2 and 4 Superclasses, and the binary Malignancy task. This approach ensures that the evaluation reflects the model's internal consistency and its ability to place lesions within the correct branch of the diagnostic tree, even if the specific subclass prediction is incorrect. We report performance using Weighted F1-Score as the primary metric for the hierarchical analysis. Given the natural class imbalance of the dataset (where common nevi outnumber rare malignancies), Weighted F1-Score provides a measure of total clinical utility, rewarding models that perform well on the most prevalent cases faced in daily practice. However, to ensure this does not mask failures on rare classes, we complement this with Balanced Accuracy in our summary tables and detailed subclass analysis.


\section{Results}

\subsection{Hierarchical Performance Benchmark}
We quantified the classification performance of the foundation models across the four levels of the clinical taxonomy. Table~\ref{tab:foundation_models} presents the "Leaderboard", the peak performance achieved by the best adapter for each foundation model. Comprehensive performance tables for all adapters are provided in Supplementary Tables S4--S13. Figure~\ref{fig:f1_box} complements this by showing the distribution of performance across all adapters via box plots.

\begin{table}[ht!]
    \centering
    \caption{Foundation Model Leaderboard. Peak Weighted F1-Score  (\%) achieved by the best-performing adapter for each foundation model across the diagnostic hierarchy. The columns represent increasing levels of clinical granularity: Malignancy (2) (Binary Benign vs. Malignant); Super (2) (Melanocytic vs. Non-Melanocytic); Super (4) (Melanocytic Benign, Melanocytic Malignant, Non-melanocytic Benign, Non-melanocytic Malignant); Main (15) (15 main diagnostic classes); and Subclass (40) (40 fine-grained lesion subtypes). Models are sorted by their performance on the finest granularity (Subclass 40). \textbf{Bold} values indicate the best performance, and \textit{italic} values indicate the second-best performance.}
    \label{tab:foundation_models}
    \makebox[\textwidth]{%
    \begin{tabular}{lcccccc}
    \toprule
           Foundation Model &  Domain &           Malignancy (2) &        Super (2) &        Super (4) &        Main (15) &    Subclass (40) \\
    \midrule
           MedSigLIP & General Medical &          \textit{96.43\%} &          \textit{93.85\%} &          \textit{91.66\%} &          \textit{62.29\%} & \textbf{69.79\%} \\
     Derm Foundation & Dermatology &          96.04\% &          93.23\% &          90.91\% &          61.34\% &          \textit{69.50\%} \\
               MONET & Dermatology &          96.02\% &          92.57\% &          89.81\% &          60.29\% &          69.31\% \\
      DINOv2 (Giant) & General Vision &          95.61\% &          92.57\% &          89.68\% &          60.06\% &          68.00\% \\
        CLIP (Large) & General Vision &          95.80\% &          92.07\% &          89.40\% &          60.41\% &          67.81\% \\
              DINOv3 & General Vision &          95.60\% &          92.61\% &          89.87\% &          60.53\% &          66.78\% \\
       DINOv2 (Base) & General Vision &          95.99\% &          91.95\% &          89.00\% &          58.44\% &          65.95\% \\
    MedImageInsights & General Medical & \textbf{97.52\%} & \textbf{95.39\%} & \textbf{93.45\%} & \textbf{62.40\%} &          65.50\% \\
         CLIP (Base) & General Vision &          95.29\% &          91.18\% &          88.12\% &          56.75\% &          64.64\% \\
      PanDerm (Base) & Dermatology &          95.59\% &          92.00\% &          89.12\% &          57.73\% &          64.12\% \\
      ResNet-50      & General Vision &         94.51\% &           89.37\% &           85.97\% &           52.88\% &           58.82\% \\
          BiomedCLIP & General Medical &          94.99\% &          89.93\% &          86.34\% &          55.00\% &          58.78\% \\
    PanDerm (Large)   & Dermatology &                91.79\%&            84.49\% &       78.67\% &          39.13\% &           36.65\% \\
    \bottomrule
    \end{tabular}
    }
\end{table}

Our hierarchical analysis exposed a significant divergence in model capabilities relative to diagnostic granularity. At the coarsest levels, the MedImageInsights model demonstrated clear superiority, achieving highest Weighted F1-scores on Binary Malignancy (97.52\%), 2-class Superclass (95.39\%), and 4-class Superclass (93.45\%) tasks. This dominance extended partially to the 15 Main Classes, where it retained a narrow lead (62.40\%). However, a distinct performance inversion occurred at the finest level of the taxonomy (40 Subclasses). MedSigLip achieved the top performance of 69.79\%, followed closely by the dermatology-specific Derm Foundation (69.50\%) and MONET (69.31\%). DINOv2 (68.00\%) outperformed MedImageInsights (65.50\%) in this fine-grained regime. Standard baselines such as ResNet-50 and BiomedCLIP trailed significantly, achieving only 58.82\% and 58.78\% respectively, underscoring the advancement of modern foundation models over previous standards.

\subsection{Qualitative Analysis of Embedding Spaces}
\begin{figure}[ht!]
    \centering
    \includegraphics[width=0.8\textwidth]{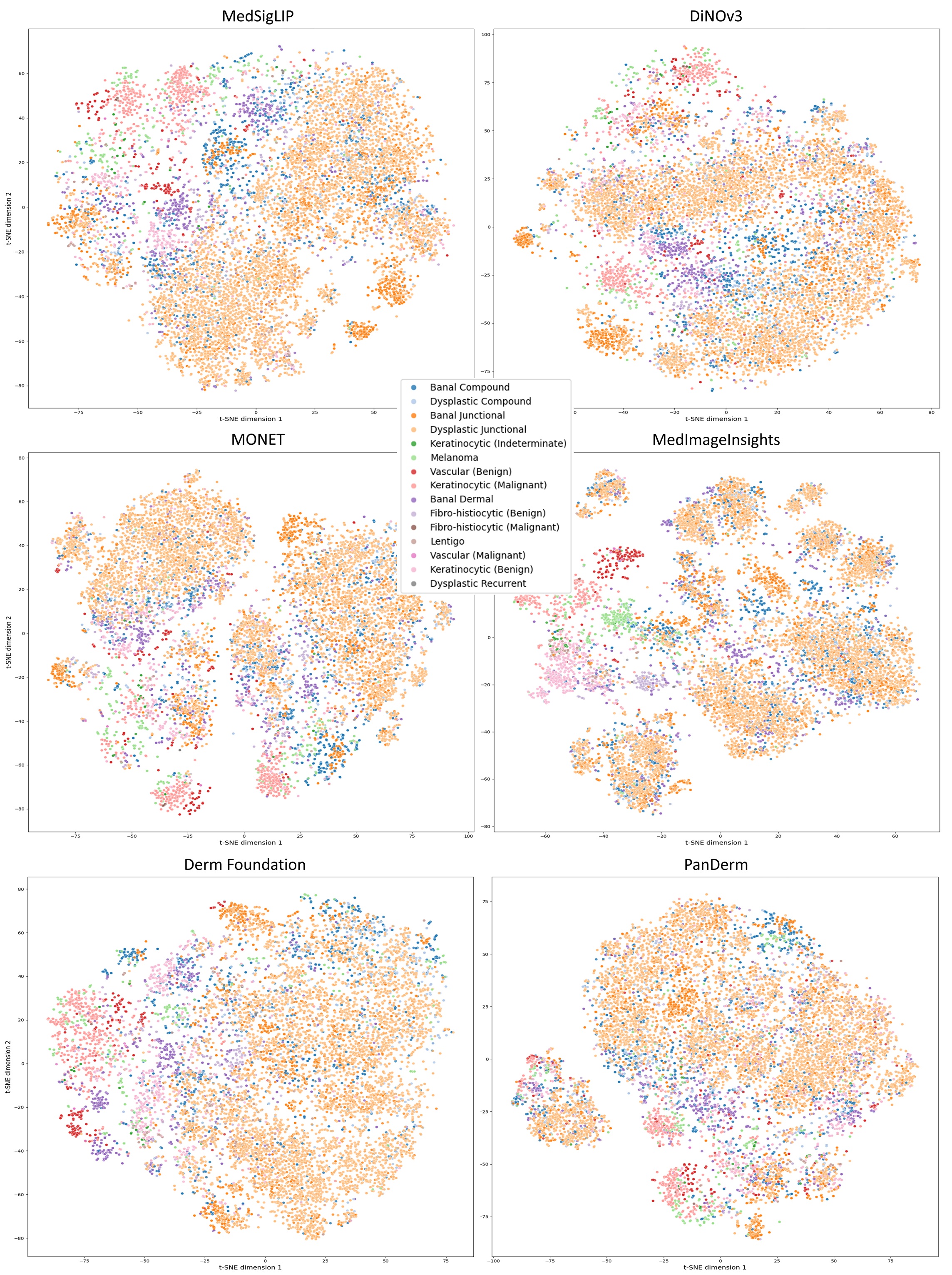}
    \caption{Projections are colored by the 15 Main Classes of the DERM12345 taxonomy. The panels display projections for MedSigLip, DINOv3, Monet, Derm Foundation, PanDerm, and MedImageInsights. Each point represents a skin lesion, colored according to the 15 Main Classes of the DERM12345 taxonomy. The plots illustrate the topological organization of the embeddings, showing the separation of distinct peripheral classes (e.g., Vascular) and the dense clustering of the central melanocytic region.}
    \label{fig:tsne_fig}
\end{figure}

To qualitatively assess the semantic structure of the learned representations prior to any supervised training, we visualized the embedding spaces using t-Distributed Stochastic Neighbor Embedding (t-SNE). Figure~\ref{fig:tsne_fig} presents a multi-panel comparison of six representative foundation models: MedSigLip, DINOv3, MONET, Derm Foundation, PanDerm, and MedImageInsights. The projections are colored according to the 15 Main Classes of the taxonomy.
Comprehensive t-SNE visualizations for all evaluated models, colored according to the 15 Main Classes, are provided in Supplementary Figures S1--S3.

\begin{figure}[ht!]
    \centering
    \includegraphics[width=0.8\textwidth]{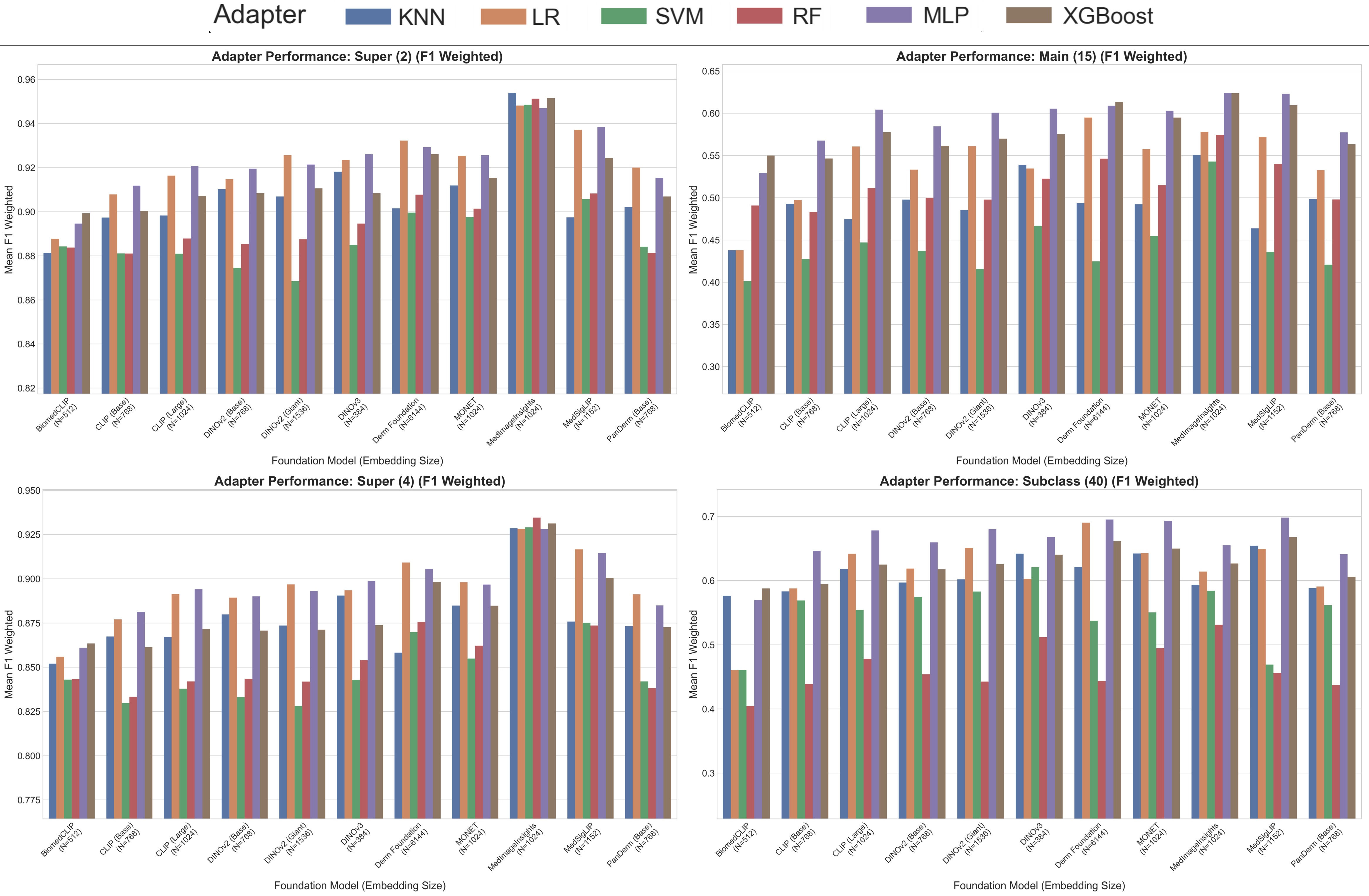}
    \caption{Hierarchical performance benchmark across four levels of diagnostic granularity. Box plots represent the distribution of Weighted F1-scores achieved by six adapter classifiers (KNN, LR, SVM, RF, MLP, XGBoost) for each foundation model. (Top Row) At the coarse Malignancy and Super Class levels, MedImageInsights achieves top performance, indicating strong semantic alignment. (Bottom Row) At the fine-grained Main Class and Subclass levels, a performance inversion occurs; MedSigLip and Monet outperform MedImageInsights, demonstrating superior capacity for granular differential diagnosis. Note the Y-axis scale differences, reflecting the increasing difficulty of the task.}
    \label{fig:f1_box}
\end{figure}

\subsection{Fine-Grained Error Analysis}
\begin{figure}[ht!]
    \centering
    \includegraphics[width=0.8\textwidth]{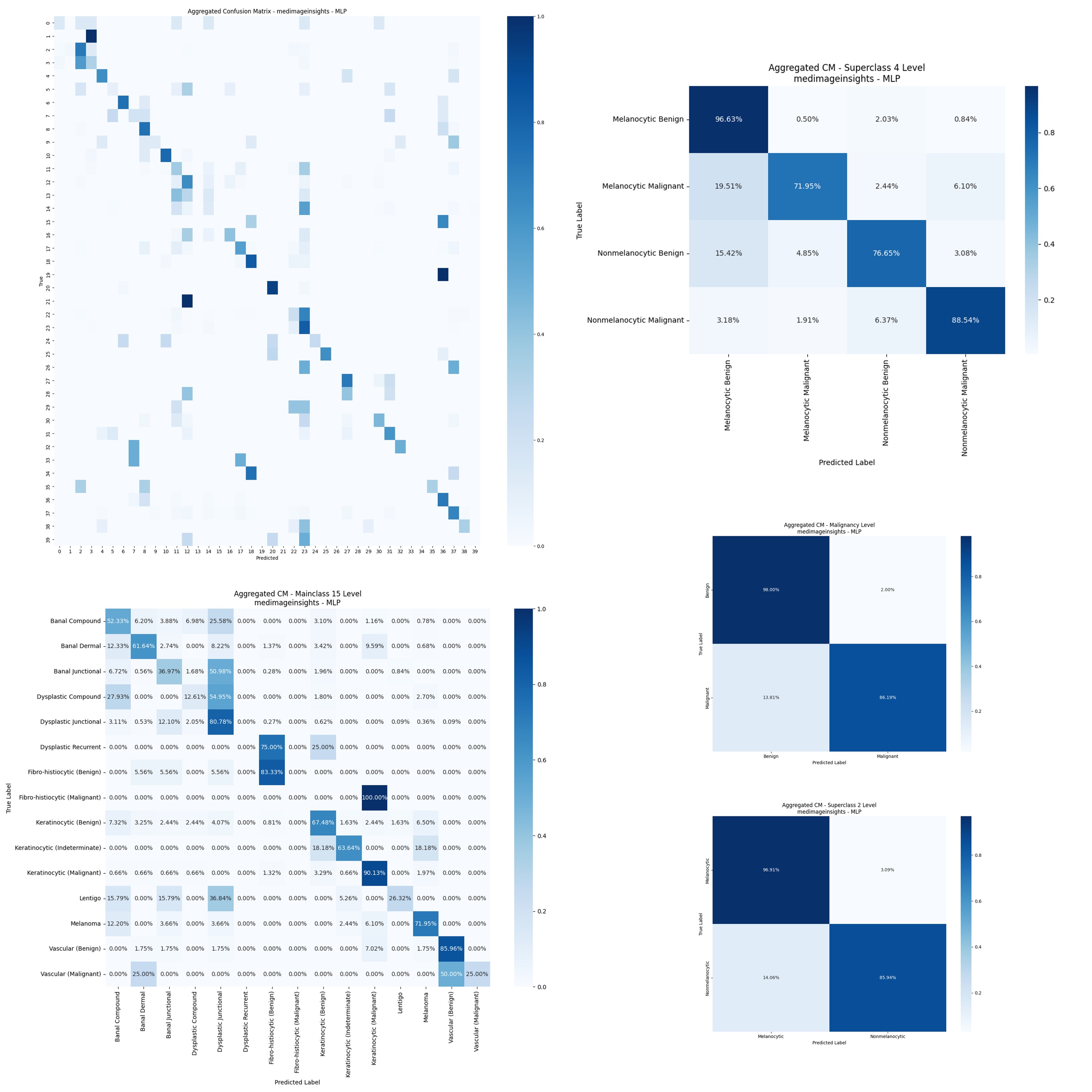}
    \caption{Multi-level confusion matrix analysis for MedImageInsights (MLP Adapter). This composite visualization illustrates the model's performance degradation across the diagnostic hierarchy. (Right Column) At the coarse levels (Malignancy, Superclass 2, Superclass 4), the model exhibits high diagonal density, effectively distinguishing broad categories like Melanocytic vs. Non-melanocytic (96.63\% accuracy for Melanocytic Benign). (Left Column) At the fine-grained levels, significant off-diagonal confusion emerges.}
    \label{fig:cm_mlp_medimageinsights}
\end{figure}

To investigate the mechanisms behind this gap, we analyzed the specific failure modes of MedImageInsights. We selected this model for detailed inspection because it represents the most illustrative case of the granularity gap: while it achieved the highest overall performance on the Binary Malignancy task (97.52\%), it suffered a sharp decline in the fine-grained Subclass task relative to visual-based models. Figure~\ref{fig:cm_mlp_medimageinsights} presents a composite confusion matrix analysis across the hierarchy. The right panels (Malignancy and Superclasses) exhibit strong block-diagonal structures, indicating high confidence in broad categorizations. For instance, the model achieves 98.00\% accuracy in correctly identifying Benign lesions (Malignancy Level).

In contrast, the left panels (Main Class and Subclass) reveal significant semantic confusion. The 15-Class Main Matrix (bottom-left) highlights the "Blob Problem" explicitly: the model frequently misclassifies Dysplastic Compound Nevi as Banal Compound Nevi (27.93\% error rate). Detailed confusion matrices for all models on the 15-class task are shown in Supplementary Figures S4--S6. Hierarchical ROC curves are presented in Supplementary Figure S7. These two classes are biologically distinct (one is a potential precursor to melanoma, the other is harmless) but visually similar. The model's inability to separate them, despite its high binary accuracy, underscores that its embeddings cluster "nevi" together semantically but fail to resolve the intra-class visual continuum required for differential diagnosis.

\section{Discussion}
The most significant finding of this benchmark is the divergence in model performance across the diagnostic hierarchy, a phenomenon we term the "granularity gap." We observed that MedImageInsights, a model trained with extensive image-caption supervision, achieved state-of-the-art performance on binary malignancy detection. This suggests that the model has effectively encoded high-level semantic concepts, learning to distinguish "cancer" from "non-cancer" based on the linguistic signals in its training data. However, this semantic strength appears to mask a deficit in fine-grained visual discrimination. On the 40-class subclass task, its performance dropped, falling behind general medical and dermatology-specific models. The mechanism of this failure is elucidated by the confusion matrix analysis, which revealed a high error rate in distinguishing Dysplastic Compound Nevi from Banal Compound Nevi. Clinically, this is a critical distinction, as dysplastic nevi serve as potential markers for melanoma risk. The model's inability to resolve these visually similar but biologically distinct entities indicates that its embedding space clusters lesions based on broad semantic categories rather than the subtle textural features required for differential diagnosis.

It is important to contextualize these findings within our specific hierarchical training strategy. Unlike standard benchmarks that might train separate models for each task (e.g., a dedicated binary classifier), we trained our adapters solely on the fine-grained 40-class labels and generated coarse-grained predictions by aggregating the probabilities of the constituent subclasses. We then calculated performance for the coarser taxonomic levels (Main Class, Superclass, and Malignancy) by explicitly mapping each of the 40 subclasses to their respective parent categories and aggregating the predicted probabilities. This approach is more detailed than direct training for every hierarchical level. It forces the adapters to learn the specific visual features of each subclass; a correct binary prediction can only be achieved if the model correctly identifies the lesion or confuses it with another lesion in the same malignancy category. Consequently, the high binary accuracy achieved by models like MedImageInsights and MedSigLip (97.52\% and 96.43\% Weighted F1, respectively) via this aggregation method is a strong validation of their internal consistency. It implies that even when these models fail to distinguish between two specific subtypes (e.g., two types of benign nevi), they correctly map the lesion to the appropriate branch of the taxonomic tree.

Our results also challenge the assumption that domain specificity is the sole determinant of performance. While the dermatology-specific Derm Foundation and MONET models performed exceptionally well, the general medical model MedSigLip achieved the top rank on the subclass task. This implies that state-of-the-art general medical models, when trained on sufficient scale and diversity, can learn feature representations that match or exceed those of specialized models. Importantly, this benchmark highlights the superiority of modern foundation models over traditional baselines. The standard ResNet-50 model, representing the previous era of supervised learning, achieved only 58.82\% accuracy on the subclass task—lagging more than 10 percentage points behind leaders like MedSigLip (69.79\%) and Derm Foundation (69.50\%). This substantial performance gap empirically validates the paradigm shift in medical image analysis, confirming that embeddings derived from large-scale pretraining provide significantly richer diagnostic signals than standard CNN architectures. The robust performance of DINOv2 and DINOv3 further supports this, demonstrating that large-scale learning on natural images can yield features that are surprisingly transferable to dermatology, provided the downstream adapter is sufficiently expressive. However, domain-specific pretraining is not always a guarantee of success, as illustrated by the varying performance within the PanDerm family\cite{Yan2025_panderm}. While PanDerm (Base) was competitive (64.12\%), the larger PanDerm (Large) model performed poorly, dropping dramatically to 36.65\% and falling behind a standard ResNet-50. Such a drastic performance drop in a larger, presumably more capable model is highly unusual and suggests that the publicly released model checkpoint might be defective or broken. This case underscores a critical takeaway: researchers cannot simply assume a specialized model is inherently superior; thorough verification against robust generalist baselines is essential.

At the coarse levels of the hierarchy, MedImageInsights demonstrates state-of-the-art capabilities, achieving a remarkable 97.52\% accuracy on the binary malignancy task and 95.39\% on the 2-class superclass task (Melanocytic vs. Non-Melanocytic). This dominance suggests that its pretraining on captioned medical images aligns perfectly with high-level diagnostic categories. However, a dramatic inversion occurs at the finer granularities. At the Subclass (40) level, the performance of MedImageInsights drops to 65.50\%. In this regime, MedSigLip (69.79\%) and Derm Foundation (69.50\%) emerge as the leaders. Our evaluation of adapter sensitivity reveals that simple linear probing is insufficient for maximizing the utility of dermatological embeddings on fine-grained tasks. Across almost all foundation models, the Multi-Layer Perceptron (MLP) consistently achieved the highest F1-scores, followed closely by Gradient Boosting (XGBoost). For instance, on the MedSigLip embeddings, the MLP adapter (69.8\%) significantly outperformed the Support Vector Machine (46.9\%). This indicates that while foundation models provide rich representations, the decision boundaries between the 40 distinct lesion subclasses are complex and non-linear. Furthermore, the performance gap illustrates the necessity of our 'best-of-suite' evaluation strategy, particularly for high-dimensional models. For Derm Foundation, which produces massive 6144-dimensional embeddings, distance-based methods like KNN lagged behind (62.1\%), likely suffering from the curse of dimensionality, whereas the MLP successfully extracted state-of-the-art performance (69.5\%). This confirms that evaluating foundation models requires a diverse adapter suite to decouple the quality of the representation from the limitations of the classifier.

The visualizations of embedding space reveal a critical challenge in dermatological modeling: the "Blob Problem." In all six projections, the core melanocytic classes; Banal Compound, Dysplastic Compound, and Dysplastic Junctional nevi form a dense, overlapping central continent. This suggests that none of these foundation models, regardless of domain pretraining, achieve clear linear separability for the most clinically challenging cases (distinguishing difficult nevi from early melanoma) without supervised adaptation.
However, distinct behaviors emerge in the peripheral clusters. MedImageInsights shows the clearest separation for distinct, visually obvious pathologies such as Vascular (red/purple lesions) and Keratinocytic classes, forming tight "islands" away from the central mass. This indicates strong semantic alignment for distinct biological entities. In contrast, MedSigLip and DINOv3 exhibit more diffuse, cloud-like structures. While less visually distinct in 2D projection, this dispersed structure may preserve richer intra-class variance, potentially aiding the fine-grained discrimination of visually similar subtypes during the supervised adaptation phase. Across all foundation models, banal and dysplastic nevi formed a dense, overlapping cluster that lacked clear separation in the unsupervised state. The sensitivity analysis of the adapters confirms this difficulty. Linear classifiers (SVM, Logistic Regression) and Multi-Layer Perceptrons consistently outperformed distance-based methods (KNN) and tree-based ensembles (Random Forest). This indicates that while the lesion subclasses are not naturally clustered in the Euclidean embedding space, they remain separable via high-dimensional hyperplanes. This finding underscores the necessity of supervised adaptation; "zero-shot" retrieval approaches are likely insufficient for safe dermatological practice given the current state of foundation model representations.

We acknowledge several limitations in this study. First, the DERM12345 dataset was collected exclusively from clinical sources in Türkiye. Consequently, the distribution of skin phototypes may not fully represent the global population, potentially affecting the generalizability of the embeddings to diverse demographics. Second, our evaluation is restricted to dermoscopic images. While this modality offers high diagnostic precision, the performance of these foundation models on clinical (macroscopic) photography remains to be verified. Future work should extend this hierarchical benchmarking framework to multi-centric datasets and diverse imaging modalities to ensure broader clinical applicability.

In conclusion, this benchmark study demonstrates that the choice of foundation model in dermatology is not a one-size-fits-all decision but depends heavily on the clinical granularity of the task. The selection of the vision encoder is critical and must be decided based on the specific use case. By evaluating embeddings across the full hierarchy, from binary malignancy to 40 distinct subclasses, we provide a roadmap for selecting the appropriate computational backbone for diverse dermatological applications, ensuring that AI tools align with the nuanced reality of clinical diagnosis.

\bibliographystyle{unsrt}
\bibliography{derm12345_embeddings}

\end{document}